\documentclass[sigconf,nonacm]{acmart}

\setcopyright{none}
\renewcommand\footnotetextcopyrightpermission[1]{}

\AtBeginDocument{%
  }

\begin{document}

\title{CircuTutor: Transforming Static Circuit Problems into Intelligent and Dynamic Tutoring}

\author{Ziyu Luo}
\orcid{0009-0000-2324-0793}
\affiliation{%
  \institution{Beijing Technology and Business University}
  \city{Beijing}
  \country{China}
}
\email{ziyu.luo@st.btbu.edu.cn}

\author{Xiaorui Ma}
\orcid{0009-0008-1241-7547}
\affiliation{%
  \institution{Beijing Technology and Business University}
  \city{Beijing}
  \country{China}}
\email{xiaorui.ma@st.btbu.edu.cn}

\author{Lin Chen}
\orcid{0009-0009-8696-1788}
\affiliation{%
  \institution{Beijing Technology and Business University}
  \city{Beijing}
  \country{China}}
\email{lin.chen@st.btbu.edu.cn}

\author{Xiaoming Chen}
\authornote{Corresponding author.}
\orcid{0009-0002-7503-3021}
\affiliation{%
  \institution{Beijing Technology and Business University}
  \city{Beijing}
  \country{China}}
\email{xiaoming.chen@btbu.edu.cn}





\renewcommand{\shortauthors}{Luo et al.}

\begin{abstract}
Learning direct current circuit concepts requires learners to connect invisible physical quantities, such as current, voltage, resistance, and power, with observable outcomes such as bulb brightness. Conventional textbook materials and general-purpose circuit simulators provide opportunities for problem solving and exploration but offer limited support for explaining why circuit behavior changes or diagnosing the reasoning behind incorrect answers. We present CircuTutor, a circuit-state-driven intelligent tutoring system that transforms static textbook circuit problems into an interactive tutoring workflow. CircuTutor first uses multimodal problem parsing to extract the textbook question, circuit topology, component parameters, switch states, and answer options, which are converted into a structured task and validated through circuit simulation. Learners can then interactively explore the circuit (by changing parameters) and submit an answer while a SPICE-compatible solver computes physically consistent circuit states. After the learner submits an answer, CircuTutor presents a before-and-after circuit state animation corresponding to the selected operation, organizes the simulated state changes into a causal reasoning chain that explains the underlying circuit behavior, maps answer discrepancies to likely misconceptions, and generates adaptive follow-up exercises targeted at the diagnosed misconception. Our experimental results demonstrate that CircuTutor effectively improves conceptual learning and the overall learning experience. The proposed framework demonstrates how simulated circuit states can be transformed into intelligent and interactive tutoring for circuit education, with the potential to generalize to other STEM domains.
\end{abstract}


\begin{teaserfigure}
  \centering
  \includegraphics[width=\textwidth]{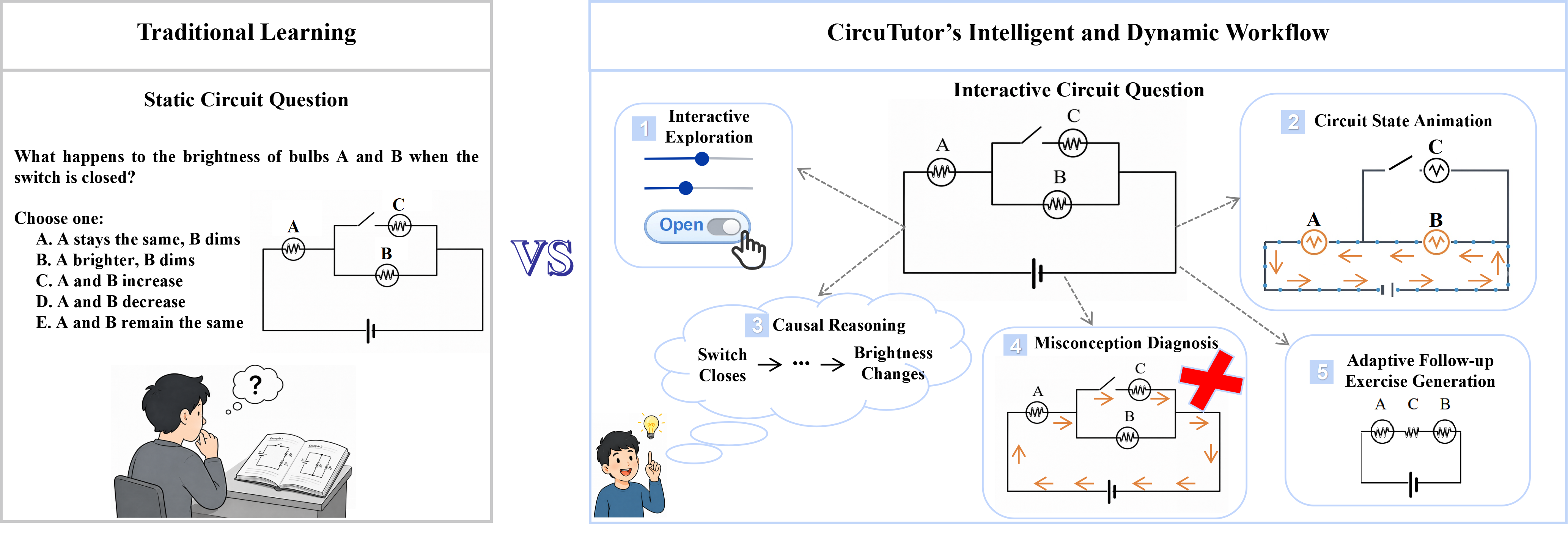}
  \caption{
  Comparison between traditional static circuit learning and the proposed CircuTutor. Traditional learning materials typically present a static circuit problem, providing limited support for understanding the reasoning behind incorrect answers. In contrast, CircuTutor transforms static circuit problems into an intelligent and dynamic tutoring workflow that integrates interactive exploration, circuit state animation, causal reasoning, misconception diagnosis, and adaptive follow-up exercise generation.
  }
  \Description{A comparison between conventional circuit learning and CircuTutor. The conventional side shows a learner studying a static circuit problem and remaining confused. The CircuTutor side shows five stages: interactive exploration and answer, physics-grounded comparison, causal reasoning, misconception diagnosis, and adaptive follow-up practice, ending with a learner who understands the circuit concept.}
  \label{fig:teaser}
\end{teaserfigure}


\maketitle

\section{Introduction}
\label{sec:introduction}

Learning direct current (DC) circuits is a fundamental yet challenging topic in physics and engineering education. To reason correctly about circuit behavior, learners must connect invisible physical quantities, including current, voltage, resistance, and power, with observable outcomes such as bulb brightness. Establishing these conceptual connections remains challenging for many learners. Previous research has shown that many learners continue to hold misconceptions after instruction, such as treating a battery as a constant-current source, confusing current with voltage or resistance, and reasoning only about local components rather than the circuit as a whole \cite{engelhardt2004students}. 
These difficulties are particularly evident in problems involving switch operations and series–parallel circuits, where a local topological change can propagate through the circuit and alter equivalent resistance, current distribution, component power, and observable brightness. 

Interactive simulations provide learners with opportunities to manipulate circuit parameters and observe the resulting behaviors. For example, research-informed platforms such as PhET \cite{perkins2006phet} support conceptual exploration by connecting scientific models with dynamic visual representations. Moreover, recent studies have demonstrated the potential of transforming static textbook content into interactive and explorable learning materials \cite{gunturu2024augmented,dai2026livephys,Dai2025PhysGest}, making existing textbook content more accessible as interactive learning experiences. However, interaction alone does not necessarily help learners explain why a circuit behaves in a particular way. General-purpose simulators \cite{dzikovska2014beetle} typically provide numerical results or visual feedback, but they rarely identify the underlying reasoning behind an incorrect answer or guide learners through the causal relationships among circuit topology, physical quantities, and observable outcomes such as brightness changes. This limitation is consistent with prior reviews of simulation-based science learning, which emphasize that simulations are most effective when integrated with appropriate instructional guidance \cite{rutten2012learning}.

Beyond enabling interactive simulation-based exploration, effective learning requires feedback that helps learners interpret simulation outcomes and revise their reasoning. Feedback is therefore a central part of the learning process, but its value depends on its specificity, timing, and explanatory content \cite{hattie2007power,shute2008focus}. In circuit learning, simply marking an answer as correct or incorrect is insufficient because the same incorrect option may reflect different misconceptions. For example, in the question in the left part of Figure~\ref{fig:teaser}, one learner may assume that adding a parallel branch makes all bulbs dimmer (i.e., D option), whereas another may recognize a local branch change but ignore its effect on the main circuit current (i.e., B option). To this end, intelligent tutoring systems have demonstrated the value of individualized feedback across learning domains \cite{ma2014intelligent}. Nevertheless, applying such support to circuit reasoning requires diagnostic feedback that is both pedagogically adaptive and grounded in physically consistent circuit states. 

To address these challenges, as illustrated in Figure~\ref{fig:teaser}, we present CircuTutor, a circuit-state-driven intelligent and dynamic tutoring system that transforms static textbook circuit problems into an interactive tutoring workflow. The workflow begins with \emph{Multimodal Problem Parsing}, which performs question parsing, answer option extraction, parameter parsing, and circuit-element and topology parsing to convert a textbook problem image into a structured circuit task. The resulting task is validated for schema completeness and simulation readiness. During \emph{Learner Interaction and Circuit Animation}, learners interactively explore the circuit and submit an answer while CircuTutor records the interaction trace and converts the learner's selected operation into a parameterized circuit model. The parameterized circuit model is compatible with SPICE, which is a widely adopted circuit simulation framework that computes electrical states \cite{nagel1975spice2,Popov2023Python}. Therefore, CircuTutor employs a SPICE-based solver to compute the corresponding voltage, current, and power, from which a before-and-after circuit state animation is generated. The resulting circuit states are passed to \emph{Misconception Diagnosis and Follow-up Exercise}, where CircuTutor constructs a causal reasoning chain, diagnoses the learner's likely misconception, and generates a follow-up exercise targeting the diagnosed misconception. The generated follow-up exercise becomes the next task in the learning cycle.



The main contributions of this work are summarized as follows.
\begin{itemize}
    \item We propose CircuTutor, a framework that transforms static textbook circuit problems into an intelligent and dynamic tutoring workflow through multimodal problem parsing and circuit simulation.
    \item We introduce a circuit-state-driven tutoring mechanism that leverages simulated circuit states to generate circuit state animation, causal reasoning, misconception diagnosis, and adaptive follow-up exercise generation.
    \item We validate the complete CircuTutor workflow through a mixed-methods controlled study, demonstrating significant improvements in conceptual understanding, transfer performance, learning efficiency, and learner experience.
\end{itemize}

Beyond circuit education, we believe that the proposed framework potentially provides a general paradigm for transforming simulation-derived state representations into intelligent and dynamic tutoring support. Rather than treating simulation as an isolated exploration tool, the proposed paradigm demonstrates how simulation outputs can be systematically translated into interpretable educational feedback, including visual explanation, causal reasoning, misconception diagnosis, and follow-up exercise generation. We anticipate that this paradigm can be extended to a wide range of simulation-based STEM learning domains, where helping learners connect invisible system states with observable phenomena is essential for conceptual understanding.

\section{Related Work}
\subsection{Interactive Simulations for Physics Education}

Interactive simulations have been widely used in physics education because they allow learners to manipulate parameters, observe system responses, and connect abstract models with observable phenomena. PhET \cite{perkins2006phet} is a representative example, providing research-informed simulations that support conceptual exploration across physics topics. Relevant research in science education suggests that simulations can improve learning when they are integrated with instructional guidance and appropriate learning activities rather than used as isolated exploratory tools \cite{smetana2012computer,de2013physical,rutten2012learning}. Such support is particularly important in DC circuit learning because current, voltage, resistance, and power are not directly visible. Learners may therefore manipulate a circuit successfully while still holding misconceptions about the relationships among these quantities \cite{engelhardt2004students}. General-purpose circuit simulators \cite{dzikovska2014beetle} typically support open-ended exploration and numerical observation, but they do not necessarily diagnose why an answer is incorrect or connect a local circuit operation to its whole-circuit consequences. CircuTutor builds on interactive simulation by embedding it within a guided tutoring loop, i.e., learners first interactively explore the circuit and commit to an answer, after which simulated circuit states are used to generate circuit state animation, causal reasoning, misconception diagnosis, and adaptive follow-up exercise generation.

\subsection{Visual Explanations and Augmented Learning Materials}

Visual explanations can help learners coordinate diagrams, numerical values, verbal descriptions, and conceptual models. Multimedia learning research emphasizes the value of integrating text and graphics \cite{mayer2002multimedia}, while research on multiple external representations highlights the need to help learners relate information distributed across different representational forms \cite{ainsworth2006deft}. In related research, several systems have further transformed static instructional materials into augmented interactive explanations. For example, PhysicsBook supports sketch-based authoring of animated physics diagrams \cite{cheema2012physicsbook}, and Augmented Math provides AR-based explorable explanations for static mathematics materials \cite{chulpongsatorn2023augmented}. Augmented Physics further uses computer vision and multimodal AI to convert textbook diagrams into embedded interactive simulations across domains including circuits, optics, and kinematics \cite{gunturu2024augmented}. The proposed CircuTutor similarly begins with a static textbook problem image, but focuses specifically on transforming DC circuit questions into validated, intelligent, and dynamic tutoring tasks. 

\subsection{Diagnostic Feedback and Intelligent Tutoring}

Feedback can support learning when it is timely, specific, and directed toward the learner's task-related reasoning rather than merely reporting correctness \cite{hattie2007power,shute2008focus}. In conceptual domains such as physics, elaborated feedback should help learners understand what was incorrect, why it was incorrect, and how their reasoning can be improved. Intelligent tutoring systems have long investigated individualized hints, learner modeling, and step-level instructional support. Meta-analytic evidence suggests that such systems can improve learning outcomes relative to conventional instructional conditions \cite{ma2014intelligent}, while VanLehn emphasizes that step-based tutoring can provide finer-grained support than answer-level feedback alone \cite{vanlehn2011relative}. Large language models create additional opportunities for generating explanations and practice materials, but their educational use also raises concerns about factual reliability and transparency \cite{kasneci2023chatgpt}. CircuTutor address these limitations by grounding circuit outcomes in reliable simulation and mapping answer discrepancies to likely misconceptions through an evidence-constrained rule set. A constrained LLM is used only to vary adaptive follow-up exercises, thereby separating reliable physical reasoning from language generation. 

\section{Method}
\label{sec:method}

\begin{figure*}[t]
  \centering
  \includegraphics[width=\linewidth]{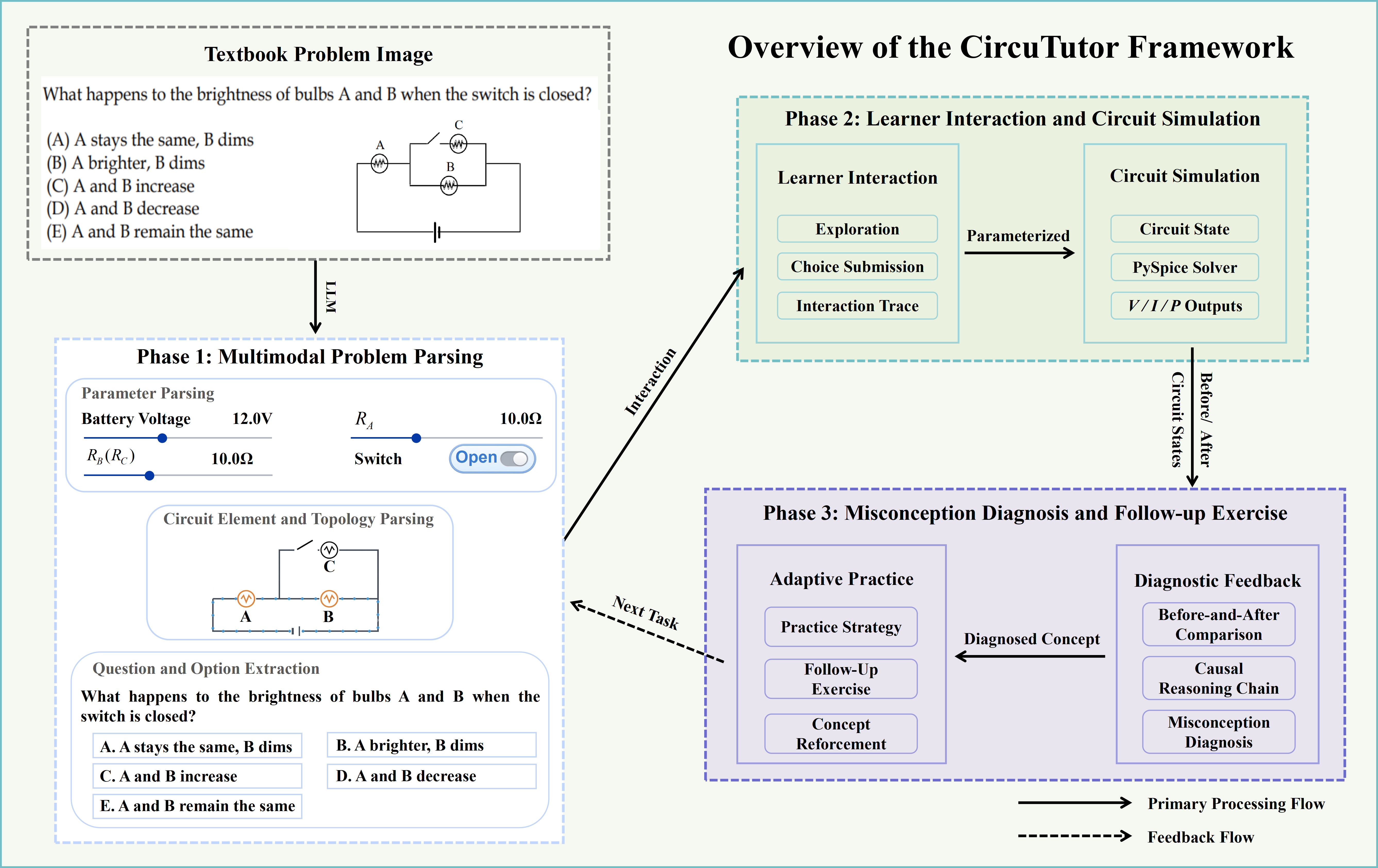}
  \caption{The three-phase architecture of CircuTutor. Phase~1 parses a textbook problem image into a structured circuit task and validates its schema completeness and simulation readiness. Phase~2 supports interactive circuit exploration, answer submission, and interaction tracing, while a PySpice-based backend constructs and solves paramterized circuit states. Phase~3 transforms the resulting circuit states corresponding to the learner's choice into a circuit state animation, causal reasoning, misconception diagnosis, and adaptive follow-up exercise generation. The circuit state animation is presented as a before-and-after comparison to illustrate the effects of the learner's selected operation. Solid arrows denote the primary processing flow, whereas dashed arrows indicate the adaptive feedback loop that returns the generated follow-up exercise as the next learning task.}
  \Description{A three-phase CircuTutor architecture. A textbook circuit image enters a multimodal parsing and task-construction phase. The validated task is passed to learner interaction and physics-grounded simulation. Simulation evidence and the learner response then enter diagnostic feedback and adaptive-practice modules. Solid arrows show the main processing direction, and dashed arrows show the feedback path that generates a new practice task.}
  \label{fig:frame}
\end{figure*}

\begin{figure*}[t]
  \centering
  \includegraphics[width=\linewidth]{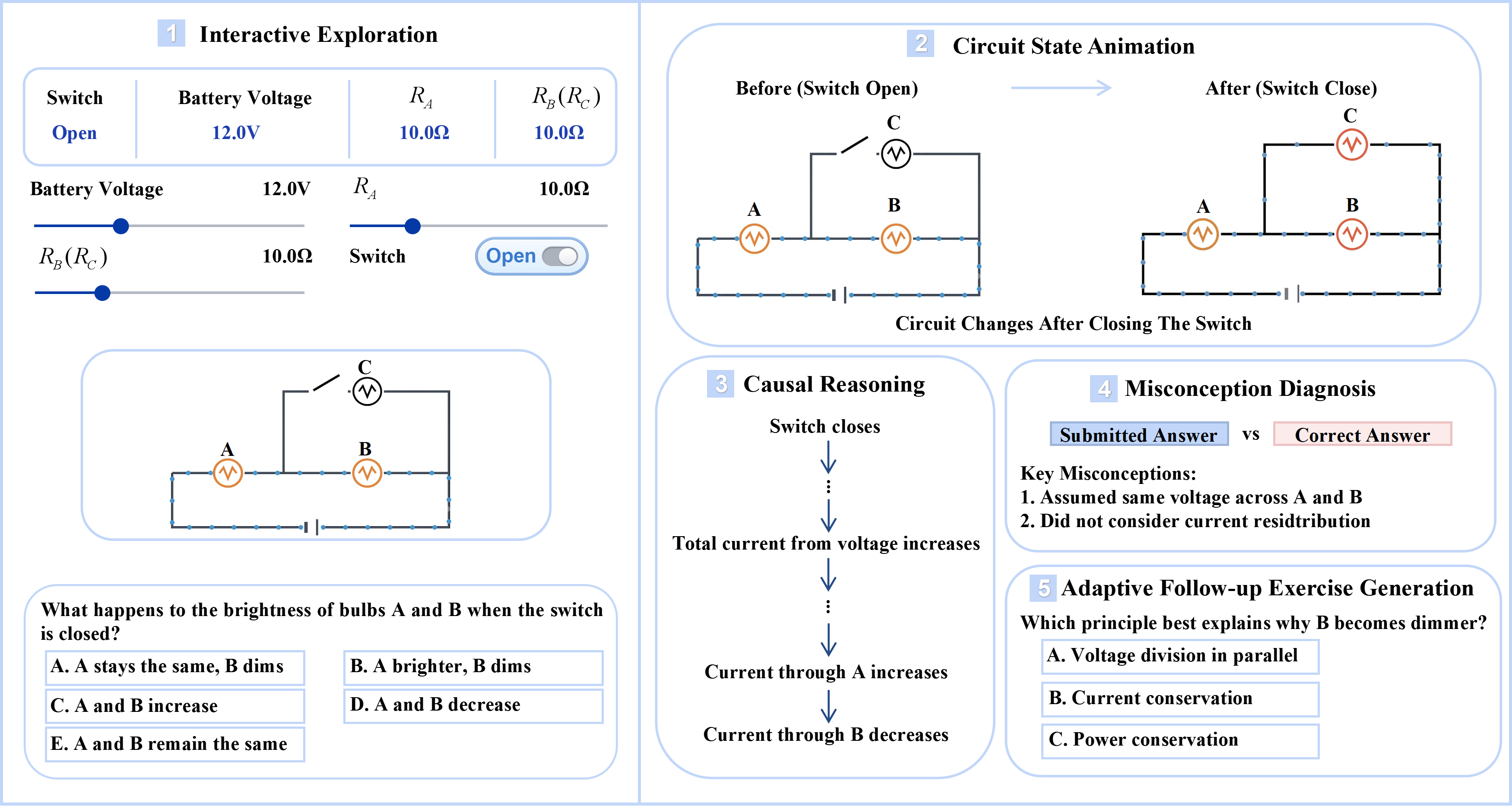}
  \caption{
  Demonstration of the learner-facing workflow of CircuTutor. In Step 1, learners interactively explore the circuit by adjusting component parameters or switch states before submitting an answer. During this stage, the interface withholds answer-revealing information such as the correct answer and explanatory feedback. In Step 2, after answer submission, CircuTutor presents a before-and-after circuit state animation to visualize the effects of the learner's selected operation. In Step 3, CircuTutor organizes the simulated state changes into a causal reasoning chain that explains the underlying circuit behavior. In Step 4, if the submitted answer is incorrect, CircuTutor identifies the most likely misconception by comparing the learner's response with the correct answer and diagnoses the concept that requires reinforcement. In Step 5, CircuTutor generates adaptive follow-up exercises tailored to reinforce the identified concept.
  }
  \Description{Screenshots of the CircuTutor interface arranged in five steps. The first step shows interactive circuit exploration and answer question. The following steps show a before-and-after comparison of physical quantities, a causal reasoning chain, a misconception diagnosis, and an adaptive follow-up exercise.}
  \label{fig:UI}
\end{figure*}

\subsection{Overview of the Learning Workflow}
\label{subsec:workflow}
CircuTutor transforms a static textbook circuit problem into intelligent and dynamic tutoring activities.
As shown in Figure~\ref{fig:frame}, CircuTutor comprises three phases, which are described in the following subsections. Phase~1, \emph{Multimodal Problem Parsing}, converts a textbook problem image into a validated structured task containing the question, answer options, circuit topology, component parameters, and visual mappings. Phase~2, \emph{Learner Interaction and Circuit Simulation}, records learner exploration, answer submission, and interaction traces, and performs SPICE-compatible \cite{nagel1975spice2,Popov2023Python} circuit simulation to compute physically consistent circuit states, which are further visualized as circuit state animation. Phase~3, \emph{Misconception Diagnosis and Follow-up Exercise}, combines the learner's submitted answer with the simulated circuit states to construct a causal reasoning chain, diagnose likely misconceptions, and generate adaptive follow-up exercises. 

Figure~\ref{fig:UI} demonstrates the corresponding learner-facing workflow. Before learners submit an answer, CircuTutor supports interactive exploration while hiding solution-related information, including bulb powers, brightness-change labels, the correct answer, and explanatory feedback. This design prevents the interface from revealing the solution before learners commit to their own reasoning.


\subsection{Multimodal Problem Parsing}
\label{subsec:parsing}

As shown in the left part of Figure~\ref{fig:frame}, CircuTutor takes as input a textbook circuit problem image containing the problem statement, answer options, a circuit diagram, component labels, electrical parameters, switch configurations, etc. The multimodal problem parsing module performs parameter parsing, circuit element extraction, and answer option extraction to convert the input into a structured circuit task. It identifies circuit elements, including voltage sources, bulbs, wires, and switches, together with their attributes, connectivity, and visual correspondences. Rather than processing these elements independently, CircuTutor organizes them into a structured circuit task ${T}$:

\begin{equation}
{T} =
\left(
{G},
{\theta},
{S},
{O},
{V}
\right),
\label{eq:task-representation}
\end{equation}

\noindent where ${G}$ denotes the circuit topology, ${\theta}$ contains component parameters, ${S}$ specifies the target state transition, ${O}$ contains the answer options and their corresponding circuit states, and ${V}$ maps simulated circuit elements to their visual representations. The same task representation is shared by the all components of CircuTutor.

CircuTutor employs SPICE (Simulation Program with Integrated Circuit Emphasis) \cite{nagel1975spice2}, a widely used circuit simulator for computing electrical states. Before a task is presented to learners, schema validation checks that its component references, parameters, answer mappings, and visual mappings are complete and internally consistent. Simulation validation then instantiates the states specified by ${S}$ and verifies that both can be solved by the PySpice solver \cite{Popov2023Python}, which is a python implementation of SPICE. Only tasks that satisfy both conditions are passed to the learner interaction and circuit simulation phase.

\subsection{Learner Interaction and Circuit Simulation}
\label{subsec:interaction-physics}

As shown in the upper-right part of Figure~\ref{fig:frame}, learners can interactively explore each circuit problem by adjusting the battery voltage, component resistances, or switch state before submitting an answer. At interaction step $t$, these operations define a parameterized state $x_t=(\boldsymbol{\theta}_t,\mathbf{s}_t)$, where $\boldsymbol{\theta}_t$ contains the current numerical parameters and $\mathbf{s}_t$ contains the switch states. Each state update is also recorded in the interaction trace. CircuTutor converts $x_t$ and the task representation ${T}$ into a SPICE-compatible circuit and invokes a PySpice-based solver. Voltage sources and resistive bulbs are represented directly, while closed and open switches are modeled as low- and high-resistance paths, respectively, so that both configurations can be solved within the same circuit formulation. The resulting circuit state $\mathbf{y}_t$ is

\begin{equation}
\mathbf{y}_t =
\Phi({T},x_t)
=
\left(
I_{\mathrm{main},t},
\left\{
{q}_{i,t}
\right\}_{i=1}^{N}
\right),
\label{eq:physical-state}
\end{equation}

\noindent where $\Phi$ denotes the SPICE-compatible solver, $I_{\mathrm{main},t}$ is the main-circuit current, and ${q}_{i,t}=(V_{i,t},I_{i,t},P_{i,t})$ contains the voltage, current, and power of component $i$, with $P_{i,t}=|V_{i,t}I_{i,t}|$. Bulb brightness is therefore derived from simulated power rather than manually assigned.
The standardized state $\mathbf{y}_t$ is shared by all post-submission components. The visual comparison is deterministically derived from the PySpice solution, whereas the causal reasoning chain and misconception diagnosis are performed using Qwen3.6 \cite{yang2025qwen3technicalreport}. Specifically, Qwen3.6 is conditioned on the structured task representation, the learner's submitted answer, and the simulation-grounded evidence derived from the same circuit solution. Consequently, although the explanatory and diagnostic content is generated by the large language model, all post-submission components remain ground in a common, physically consistent circuit state rather than being produced as unconstrained and independently generated descriptions.

\subsection{Misconception Diagnosis and Follow-up Exercise}
\label{subsec:diagnosis-practice}

As shown in the lower-right part of Figure~\ref{fig:frame} and Step~1 of Figure~\ref{fig:UI}, learners can interactively explore the circuit by adjusting component parameters or switch states before submitting an answer. After learners submit their answers, CircuTutor computes the circuit states before and after the target operation, $\mathbf{y}^{-}$ and $\mathbf{y}^{+}$. Their difference is summarized as simulation-grounded evidence $\mathbf{e}=(\Delta I_{\mathrm{main}},\{\Delta V_i,\Delta I_i,\Delta P_i,c_i\}_{i=1}^{N})$, where each difference is computed as the after-state value minus the before-state value. The categorical outcome $c_i\in\{-1,0,+1\}$ represents whether bulb $i$ becomes dimmer, remains unchanged, or becomes brighter. It is determined from $\Delta P_i$ using a small numerical tolerance so that negligible solver-level differences are not presented as meaningful brightness changes. The evidence $\mathbf{e}$ drives the before-and-after comparison component in Figure~\ref{fig:frame}, which is presented to learners as the circuit state animation in Step~2 of Figure~\ref{fig:UI}. Through the visual mapping ${V}$, each component-level voltage, current, power, and brightness change is attached to the corresponding element in the rendered circuit. This allows learners to inspect both the structural operation and its physical consequences. 

CircuTutor then instantiates a causal reasoning chain from an ordered physical-dependency template. For switch-induced circuit changes, the chain begins with the switch operation and the resulting topology change, and then traces how this change affects the equivalent resistance, main-circuit current, component-level voltage and current, electrical power, and finally the observable brightness variation. The active topology transition determines which steps are included, while the simulation result supplies the corresponding directions of change. The causal reasoning chain provides learners with an explanation of how a circuit operation propagates through intermediate physical quantities to produce the observed brightness variation. This internally generated chain is presented in Step~3 of Figure ~\ref{fig:UI}. 

For misconception diagnosis, which is presented in Step~4 of Figure~\ref{fig:UI}, each answer option is mapped to a predicted bulb-change pattern $\hat{\mathbf{c}}=(\hat{c}_1,\ldots,\hat{c}_N)$, while the physical evidence provides the corresponding simulated pattern $\mathbf{c}=(c_1,\ldots,c_N)$. A rule-based matcher identifies which parts of the learner's answer conflict with the simulated outcome and selects the first evidence-consistent misconception rule: 

\begin{equation}
m^{*} =
\operatorname{first}
\left\{
m\in{M}_{\mathrm{mis}}:
r_m(\hat{\mathbf{c}},\mathbf{c},
\mathbf{e},{T})=1
\right\},
\label{eq:misconception-mapping}
\end{equation}

\noindent where ${M}_{\mathrm{mis}}$ is the ordered set of misconception categories and $r_m$ is the evidence condition associated with category $m$. If no category-specific rule is satisfied, the system returns a generic discrepancy explanation rather than assigning an unsupported misconception. The current rule set covers series-parallel confusion, constant-brightness bias, locality-based reasoning, and confusion between main current and branch power \cite{engelhardt2004students}.

Finally, the diagnosed misconception $m^{*}$ is used to determine the concept that requires reinforcement with a selection of a follow-up strategy, including a contrastive circuit variant, parameter-exploration task, or multiple-choice explanation question. The resulting follow-up exercise generation is presented in Step~5 of Figure~\ref{fig:UI}. It preserves the target physical dependency while varying the circuit configuration, parameters, or response form. When language variation is needed, a constrained LLM receives the diagnosis, task constraints, and
simulation evidence; however, it does not determine the correct physical outcome. The generated task is checked against the same 
physics-grounding process before being returned to the next learning cycle \cite{kasneci2023chatgpt}. 

\section{Evaluation}
\label{sec:evaluation}

\subsection{Overview}
\label{subsec:evaluation-design}

We conducted two complementary evaluations of CircuTutor. First, a between-subjects controlled evaluation assessed the learning outcomes and learner experience by comparing CircuTutor with a control condition. Second, an evaluation of tutoring support examined the quality of CircuTutor's support through expert review and learner feedback. Therefore, these evaluations examine not only the overall learning effectiveness and experience of CircuTutor but also the quality of its tutoring support.

\subsection{Evaluation of Learning Outcomes and Experience}
\label{subsec:participants-procedure}

\subsubsection{Evaluation Design} 
\hfill\break
\noindent\textbf{Participants.} 
The participant sample consisted of 30 students who had previously received introductory instruction on DC circuits but had not used CircuTutor. The sample included 12 male and 18 female participants, with a mean age of 23.97 years ($SD=2.33$). Participants were randomly assigned to either the control group ($n=15$) or the CircuTutor group ($n=15$). Before participation, all participants provided informed consent and completed a brief background questionnaire covering their prior physics coursework and experience with circuit simulation tools. Participants who did not complete all study stages or whose interaction records were incomplete were excluded from the final analysis. 

\noindent\textbf{Conditions and learning tasks.} 
Both groups completed the same four DC circuit reasoning tasks in the same order and within the same overall time budget. The tasks covered series-parallel structures, switch-induced topology changes, equivalent resistance, current redistribution, bulb power, and brightness variation. They used different circuit configurations and parameter values but required learners to coordinate the same underlying physical quantities emphasized in Section~\ref{sec:method}. The control group received a concise learning handout, written answer explanations, and access to a conventional circuit simulator. The CircuTutor group followed the learning workflow described in Section~\ref{sec:method}. All the four tasks were produced through the multimodal problem parsing and task-construction pipeline described in Section~\ref{sec:method}. They were manually verified for textual accuracy, topological correctness, and simulation validity before the study. Both groups received the same question statements, circuit configurations, answer options, task order, and time budget. 

\noindent\textbf{Procedure and logged data.} 
The study followed the same main sequence for both groups: pre-test, condition-specific tutorial, learning tasks, immediate post-test, transfer test, and questionnaire. During the learning stage, participants completed four circuit reasoning tasks under their assigned condition. Adaptive follow-up exercises for the CircuTutor group were completed within the fixed learning period so that both groups had the same overall time budget. Participants in the CircuTutor condition additionally rated the four post-submission tutoring components and completed a short interview about the generated support. The complete study procedure is summarized in Table~\ref{tab:procedure}.

\begin{table}[h]
\caption{The procedure for the evaluation of learning outcomes and experience.}
\label{tab:procedure}
\centering
\small
\begin{tabular}{lll}
\toprule
\textbf{Stage} & \textbf{Time} & \textbf{Activity} \\
\midrule
Pre-test      & [10 min] & Prior circuit knowledge \\
Tutorial      & [5 min]  & Tool instructions \\
Learning      & [20 min] & Four circuit tasks \\
Post-test     & [8 min]  & Conceptual understanding \\
Transfer test & [7 min]  & Unseen circuit problems \\
Questionnaire & [5 min]  & SUS, Paas, and IMI \\
Interview     & [10 min] & CircuTutor group only \\
\bottomrule
\end{tabular}
\end{table}

During the learning stage, CircuTutor recorded first-attempt correctness, total attempts, task-completion time, parameter adjustments, switch operations, and follow-up exercise completion. Only the outcome and efficiency measures defined in Section~\ref{subsec:measures} were used for the between-condition analysis. 

\subsubsection{Measures} 
\label{subsec:measures}
\hfill\break
\noindent\textbf{Measures for Learning outcomes.} 
Conceptual understanding was assessed using the pre-test and its parallel post-test. Scores were converted to percentages, and raw learning gain was calculated as the post-test score minus the pre-test score. Transfer test performance was measured using unseen circuit configurations and parameter combinations that required learners to apply the same principles to new problems. Post-test and transfer-test scores were treated as the primary learning outcomes, while learning gain was reported as a supplementary measure.  

\noindent\textbf{Measures for Experience.} 
For user interaction, the interaction logs provided three process measures across the learning tasks: first-attempt accuracy, task-completion time, and retry count. First-attempt accuracy was the proportion of tasks answered
correctly on the initial submission. Task-completion time was measured from task presentation to completion of the main question, excluding the tutorial and adaptive follow-up exercise. Retry count denoted the number of additional submissions after the first attempt. Because completion times were positively skewed, inferential analysis was performed on log-transformed values, while the table reports the original means and standard deviations for interpretability. 
After each learning task, participants rated their invested mental effort using the single-item nine-point Paas scale \cite{paas1992training}; ratings were averaged across the four tasks. After the learning session, both groups completed the interest/enjoyment and value/usefulness subscales of the Intrinsic Motivation Inventory (IMI) \cite{mcauley1989psychometric}. The ten-item System Usability Scale (SUS) was administered only to the CircuTutor group and converted to its standard 0-100 score \cite{brooke1996sus}. Because the shortened IMI subscales showed limited internal consistency in this pilot sample, their results are interpreted as exploratory subjective evidence rather than primary learning outcomes. 

\subsubsection{Results} 
\label{subsec:quantitative-results}
\hfill\break
Descriptive statistics are reported as means and standard deviations. Post-test and transfer-test scores were analyzed using analysis of covariance (ANCOVA) \cite{vanbreukelen2006ancova}, with condition as the between-subjects factor and
pre-test score as the covariate. Task-level first-attempt correctness was analyzed using mixed-effects logistic regression, with condition as a fixed effect and participant and task as random intercepts.Task-completion time was log-transformed and analyzed using a linear mixed-effects model, while retry counts were analyzed using a negative-binomial mixed-effects model. Paas and IMI scores were compared using two-sided Welch's $t$-tests. SUS scores were summarized descriptively for the CircuTutor group. Effect sizes are reported alongside $p$-values, using a significance level of $\alpha=.05$. Holm correction was applied within families of secondary outcome comparisons. 

As shown in Table~\ref{tab:quantitative-results}, the control group had numerically higher pre-test scores than the CircuTutor group, but the difference was not statistically significant ($p=.170$). After controlling for pre-test performance, the CircuTutor group achieved significantly higher post-test scores than the control group, $F(1,27)=5.89$, $p=.022$, $\eta_p^2=.18$. The CircuTutor group also showed a positive learning gain, whereas the control group showed a negative mean gain, $d=1.04$, $p=.008$. No significant between-condition difference was observed for transfer performance, $F(1,27)=0.21$, $p=.650$, $\eta_p^2=.01$. Compared with the control group, CircuTutor participants demonstrated higher first-attempt accuracy ($OR=3.64$, $p=.012$) and shorter task-completion time ($ratio=0.60$, $p=.004$). The difference in retry count was not statistically significant ($IRR=0.71$, $p=.126$). CircuTutor participants also reported lower mental effort and higher 
interest/enjoyment and value/usefulness than participants in the control group. The mean SUS score for CircuTutor was 73.33 ($SD=14.57$). 

\begin{table}[h]
\caption{Quantitative results of evaluation of learning outcomes and experience. Values are reported as mean (standard deviation).}
\label{tab:quantitative-results}
\centering
\footnotesize
\setlength{\tabcolsep}{3.2pt}
\renewcommand{\arraystretch}{0.95}
\begin{tabular}{@{}lcccc@{}}
\toprule
\textbf{Measure} &
\textbf{Control} &
\textbf{CircuTutor} &
\textbf{Effect size} &
\textbf{$p$} \\
\midrule

Pre-test (\%) &
73.33 (16.43) &
62.22 (25.56) &
$d=-0.52$ &
.170 \\

Post-test (\%) &
58.89 (25.87) &
74.44 (8.61) &
$\eta_p^2=.18$ &
.022 \\

Learning gain (pp) &
-14.44 (27.36) &
12.22 (23.96) &
$d=1.04$ &
.008 \\

Transfer test (\%) &
75.00 (28.35) &
71.67 (12.91) &
$\eta_p^2=.01$ &
.650 \\

First-attempt accuracy (\%) &
35.00 (15.81) &
60.00 (29.58) &
$OR=3.64$ &
.012 \\

Completion time (s/task) &
59.77 (42.78) &
33.57 (21.08) &
$ratio=0.60$ &
.004 \\

Retries (per task) &
0.98 (0.32) &
0.70 (0.54) &
$IRR=0.71$ &
.126 \\

Paas mental effort &
6.70 (1.15) &
3.23 (1.60) &
$d=-2.49$ &
$<.001$ \\

IMI interest/enjoyment &
2.56 (0.84) &
5.67 (0.60) &
$d=4.25$ &
$<.001$ \\

IMI value/usefulness &
4.44 (0.41) &
5.82 (0.58) &
$d=2.75$ &
$<.001$ \\

SUS &
-- &
73.33 (14.57) &
-- &
-- \\

\bottomrule
\end{tabular}
\end{table}

\subsection{Evaluation of Tutoring Support}
\label{subsec:support-evaluation}

To complement the previous evaluation, we have also evaluated whether the tutoring support generated by CircuTutor was physically reliable, diagnostically appropriate, and understandable to learners. This evaluation combined an expert review of representative system outputs with component ratings and interview feedback from participants in the CircuTutor group. 

\subsubsection{Design and Measures}
\hfill\break
\noindent\textbf{Expert review.} 
We invited three experts with backgrounds in physics education to independently evaluate 24 representative cases of CircuTutor's tutoring support. The cases were balanced across the misconception categories described in Section~\ref{subsec:diagnosis-practice}: 
series-parallel confusion, constant-brightness bias, locality-based reasoning, and confusion between main current and branch power. Each case contained the original circuit problem, an incorrect learner answer, the simulation-grounded outcome, the circuit state animation, the causal reasoning chain, the misconception diagnosis, and the adaptive follow-up exercise. Experts rated each case on a five-point scale for physical correctness, diagnostic accuracy, clarity, causal completeness, and remedial relevance. We report the mean, standard deviation, and percentage of ratings equal to or above four for each dimension. 

\noindent\textbf{Learner feedback.} 
After completing the common study procedure, participants in the CircuTutor condition rated four statements on a five-point scale. The statements assessed whether the circuit state animation helped them understand the brightness changes, whether causal reasoning supported step-by-step understanding, whether misconception diagnosis helped them recognize their error, and whether the adaptive follow-up exercise helped them correct similar reasoning. Participants also provided feedback about excessive information, unclear terminology, and desired interface improvements during the follow-up interview. The responses were analyzed using a hybrid deductive-inductive thematic approach \cite{braun2006using}. The initial coding categories reflected support from physical evidence, usefulness of causal reasoning, misconception awareness, usefulness of follow-up practice, and information load. Additional categories were created when responses revealed issues outside these initial themes. 

\begin{table}[h]
\caption{Expert review and learner feedback of CircuTutor.}
\label{tab:expert-learner-ratings}
\centering
\footnotesize
\setlength{\tabcolsep}{3.5pt}
\renewcommand{\arraystretch}{0.95}
\begin{tabular}{@{}p{0.55\columnwidth}cc@{}}
\toprule
\textbf{Measure} &
\textbf{M (SD)} &
\textbf{$\geq 4$ (\%)} \\
\midrule

\multicolumn{3}{@{}l}{\textit{Expert review}} \\

Physical correctness &
4.46 (0.63) &
93.1 \\

Diagnostic accuracy &
3.92 (0.71) &
73.6 \\

Clarity &
3.54 (0.73) &
51.4 \\

Causal completeness &
3.53 (0.93) &
52.8 \\

Remedial relevance &
3.05 (1.12) &
42.9 \\

\midrule

\multicolumn{3}{@{}l}{\textit{Learner feedback}} \\

Circuit state animation &
3.73 (0.80) &
66.7 \\

Causal reasoning &
3.13 (1.13) &
33.3 \\

Misconception diagnosis &
3.80 (0.77) &
73.3 \\

Adaptive follow-up exercise &
3.87 (1.06) &
53.3 \\

\bottomrule
\end{tabular}
\end{table}

\subsubsection{Results} 
\hfill\break
As shown in Table~\ref{tab:expert-learner-ratings}, experts rated the physical correctness of the generated support highest ($M=4.46$, $SD=0.63$), with 93.1\% of ratings equal to or above four. Diagnostic accuracy received the second-highest expert rating ($M=3.92$, $SD=0.71$), with 73.6\% of ratings equal to or above four. Clarity and causal completeness received mean ratings of 3.54 and 3.53, respectively, while remedial relevance received the lowest expert rating ($M=3.05$, $SD=1.12$). Learners gave the highest mean rating to the adaptive follow-up exercise ($M=3.87$, $SD=1.06$), followed by misconception diagnosis ($M=3.80$, $SD=0.77$) and circuit state animation ($M=3.73$, $SD=0.80$). Causal reasoning received the lowest learner rating ($M=3.13$, $SD=1.13$). The above feedback revealed four recurring observations. Participants reported that changes in current and power helped them justify the observed brightness variation, and that causal reasoning helped connect local switch operations with whole-circuit effects. They also indicated that misconception diagnosis made their original reasoning errors more explicit and that the adaptive follow-up exercise supported application to a related circuit. The main reported difficulty concerned the presentation and organization of causal 
reasoning, suggesting that its steps and terminology should be simplified and presented progressively to reduce information load. 

\section{Conclusion}
\label{sec:conclusion}
We presented CircuTutor, a circuit-state-driven interactive tutoring system that transforms static textbook circuit problems into an intelligent and dynamic tutoring workflow. CircuTutor integrates multimodal problem parsing and task validation, interactive exploration, SPICE-compatible circuit simulation, circuit state animation, causal reasoning, misconception diagnosis, and adaptive follow-up exercise generation into a unified learning process. By separating learner exploration and answer submission from post-submission feedback, CircuTutor moves beyond conventional answer checking to help learners understand how circuit behavior changes and why their answers may be incorrect.
Experimental results demonstrated that participants using CircuTutor achieved significantly higher post-test and transfer performance than those in the control group while reporting lower mental effort and higher learning motivation. Expert and learner evaluations further showed that the generated feedback was highly rated for physical correctness and diagnostic accuracy, with misconception diagnosis and adaptive follow-up exercises receiving the highest learner ratings.
Future work will investigate larger and more diverse learner populations, additional circuit structures and misconception categories, and longer-term knowledge retention. We also plan to conduct component-level ablation studies to better understand how individual feedback components contribute to learning outcomes.

\bibliographystyle{ACM-Reference-Format}
\bibliography{sample-base}


\end{document}